\documentclass[10pt,twocolumn,letterpaper]{article}

\usepackage{wacv}              

\usepackage{graphicx}
\usepackage{amsmath}
\usepackage{amssymb}
\usepackage{booktabs}
\usepackage[symbol]{footmisc}
\usepackage{multirow}
\usepackage{diagbox}

\usepackage[pagebackref,breaklinks,colorlinks]{hyperref}
\usepackage[accsupp]{axessibility} 

\usepackage[capitalize]{cleveref}
\crefname{section}{Sec.}{Secs.}
\Crefname{section}{Section}{Sections}
\Crefname{table}{Table}{Tables}
\crefname{table}{Tab.}{Tabs.}

\def\wacvPaperID{2174} 
\def\confName{WACV}
\def\confYear{2025}

\begin{document}

\title{AMNCutter: Affinity-Attention-Guided Multi-View Normalized Cutter for Unsupervised Surgical Instrument Segmentation}

\author{
{Mingyu Sheng}\\
{The University of Sydney}\\
{Sydney, NSW, Australia}\\
{\tt\small {mshe0136@uni.sydney.edu.au}}
\and
{Jianan Fan}\\
{The University of Sydney}\\
{Sydney, NSW, Australia}\\
{\tt\small {jfan6480@uni.sydney.edu.au}}
\and
{Dongnan Liu}\\
{The University of Sydney}\\
{Sydney, NSW, Australia}\\
{\tt\small {dongnan.liu@sydney.edu.au}}
\and
{Ron Kikinis}\\
{Harvard Medical School}\\
{Boston, MA, USA}\\
{\tt\small {kikinis@bwh.harvard.edu}}
\and
{Weidong Cai}\\
{The University of Sydney}\\
{Sydney, NSW, Australia}\\
{\tt\small {tom.cai@sydney.edu.au}}
}

\maketitle

\begin{abstract}
    Surgical instrument segmentation (SIS) is pivotal for robotic-assisted minimally invasive surgery, assisting surgeons by identifying surgical instruments in endoscopic video frames.
    Recent unsupervised surgical instrument segmentation (USIS) methods primarily rely on pseudo-labels derived from low-level features such as color and optical flow, but these methods show limited effectiveness and generalizability in complex and unseen endoscopic scenarios. 
    In this work, we propose a label-free unsupervised model featuring a novel module named Multi-View Normalized Cutter (m-NCutter). Different from previous USIS works, our model is trained using a graph-cutting loss function that leverages patch affinities for supervision, eliminating the need for pseudo-labels. The framework adaptively determines which affinities from which levels should be prioritized. Therefore, the low- and high-level features and their affinities are effectively integrated to train a label-free unsupervised model, showing superior effectiveness and generalization ability.
    We conduct comprehensive experiments across multiple SIS datasets to validate our approach's state-of-the-art (SOTA) performance, robustness, and exceptional potential as a pre-trained model. Our code is released at \href{https://github.com/MingyuShengSMY/AMNCutter}{https://github.com/MingyuShengSMY/AMNCutter}.
\end{abstract}


\section{Introduction}  
\label{Introduction}
    Minimally invasive surgery (MIS) has greatly improved patient experiences, including reduced pain, lower risk of infection, and shorter hospitalization period, owing to the smaller incisions compared to common open surgery \cite{maier2017surgical, maier2022surgical}. Despite its advantages, surgeons face new challenges due to the complex in vivo scenes and low-quality endoscopic frames, which are adversely affected by misting, noise, and a narrow field of view \cite{rueckert2024corrigendum}. To address these challenges and support surgeons, robotic assistance has been applied to a range of tasks, such as surgical workflow segmentation  
    and surgical instrument segmentation \cite{CascadeYue, Shah97843996GLSFormer, Yue_Zhang_Hu_Xia_Luo_Wang_2024, Paranjape978303166958AdaptiveSAM, zhao2021anchor}.

    Surgical instrument segmentation (SIS) aids surgeons by displaying distinguishable object masks for every surgical video frame. The SIS technique has evolved from traditional machine learning methods, such as Support Vector Machines, to advanced deep learning approaches, including CNNs and Transformers \cite{bouget2015detecting, RIEKE201682, laina2017concurrent, shvets2018automatic, LSKANet_Liu}. However, the supervised SIS methods require extensive annotated data, which is labor-intensive, time-consuming, and highly expertise-demanding. This leads to restrictive data diversity, which hinders model performance on unseen datasets and real-world applications, resulting in weak robustness. In contrast, unsupervised learning allows the model to be trained without manual annotations and to serve as a pre-trained model for other downstream tasks \cite{Xie10204211MAESTER, he2022masked, Fan10378316Taxonomy, Fan2024Learning, fan2024revisitingadaptivecellularrecognition, Liu9156759Unsupervised, Liu9195030PDAM}. Therefore, developing unsupervised SIS methods is a significant and valuable open problem in the SIS field.

    In recent years, unsupervised surgical instrument segmentation (USIS) has been preliminarily explored in several studies \cite{da2019self, liu2020unsupervised, sestini2023fun}, with the pseudo-label technique playing a crucial role in transforming the unsupervised task into a pseudo-supervised one. 
    In SOTA USIS studies, pseudo-labels are primarily derived from low-level image information (e.g., color and optical flow). These methods face several limitations, such as reduced robustness, limited effectiveness in complex surgical scenes, and an inability to handle multi-class segmentation. This is because pseudo-labels handcrafted from low-level features like color and optical flow sometimes struggle to differentiate between surgical instruments and patients' tissues, especially as the complexity of endoscopic video frames increases. For example, a tissue wall with intense light reflection may appear similar in color to metal surgical instruments, and incised tissues that move with the instruments may make optical flow inadequate for distinguishing them \cite{wu2024rethinkinglowqualityopticalflow}. 
    For label-free methods in the field of general image segmentation, feature extraction followed by pixel- or patch-wise clustering is a prevalent strategy \cite{melas2022deep, wang2023cut, simeoni2021localizing, hamilton2022unsupervised}. However, this approach is non-end-to-end, low-parameterized, and overly focused on the deepest features. It leads to relatively weak learning ability due to separate optimization and overlooks meaningful low-level object information (e.g., shape, color, and texture) presented in shallower feature maps \cite{Liu10607955Rethinking}, which drives our work. In addition, the limited exploration of label-free approaches in the USIS domain further motivates this study.


    Inspired by the recent graph-cutting-based segmentation methods \cite{melas2022deep, wang2023cut, deng2023learning, sheng2024revisitingsurgicalinstrumentsegmentation}, we propose a fully label-free USIS method named \textbf{A}ffinity-Attention-Guided \textbf{M}ulti-View \textbf{N}ormalized \textbf{Cutter} (\textbf{AMNCutter}). Diverging from the non-end-to-end methods that rely solely on the deepest feature map extracted from an image encoder, our approach fully parameterizes the graph-cutting process and incorporates multi-level patch affinities through a novel module termed \textit{Multi-View Normalized Cutter (m-NCutter)}. This multi-level/view module captures the semantic information and affinities from different levels, such as color sameness, shape likeness, and function proximity. Additionally, we introduce a novel attention block named \textit{Multi-View Self-Attention}, which adaptively assigns attention scores to affinities, guiding the model in prioritizing certain affinities at specific levels. For example, the model can dynamically determine whether the color affinity (low-level) or function similarity (high-level) is more crucial for segmentation. 
    
    In contrast to existing pseudo-label-based USIS works \cite{da2019self, liu2020unsupervised, sestini2023fun} and label-free segmentation studies like \cite{melas2022deep, wang2023cut, simeoni2021localizing, hamilton2022unsupervised, sheng2024revisitingsurgicalinstrumentsegmentation}, our method does not rely on pre-crafted pseudo-labels and operates within an end-to-end framework, demonstrating superior performance, robustness, and real-time capability, which are validated through comprehensive experiments across diverse datasets. In addition, its potential for use as a pre-trained model in downstream tasks has been confirmed through transfer learning experiments.

    Our primary contributions are summarized as follows:
    \begin{itemize}
        \item We propose an end-to-end USIS method called \textbf{AMNCutter}, requiring neither ground truth nor pseudo-labels for supervision. 

        \item We devise a novel \textit{Multi-View Normalized Cutter (m-NCutter)} optimized by a graph-cutting loss \textit{Normalized Cut Loss (NCut Loss)}. This architecture allows the model to be trained without any labels, perform prediction in an end-to-end manner, and fully consider the features and their affinities across various levels.


        \item We conduct extensive experiments across various datasets to validate our method's outstanding performance, robustness, and potential for use as a pre-trained model in supervised segmentation tasks.
    \end{itemize}

    In \cref{RelatedWork}, we review related SIS studies. \cref{Method} details our approach. The experimental results and their analysis are presented in \cref{ExperimentDiscussion}, followed by conclusions in \cref{Conclusion}.

\begin{figure*}[htbp]
  \centering
   \includegraphics[width=0.9\linewidth]{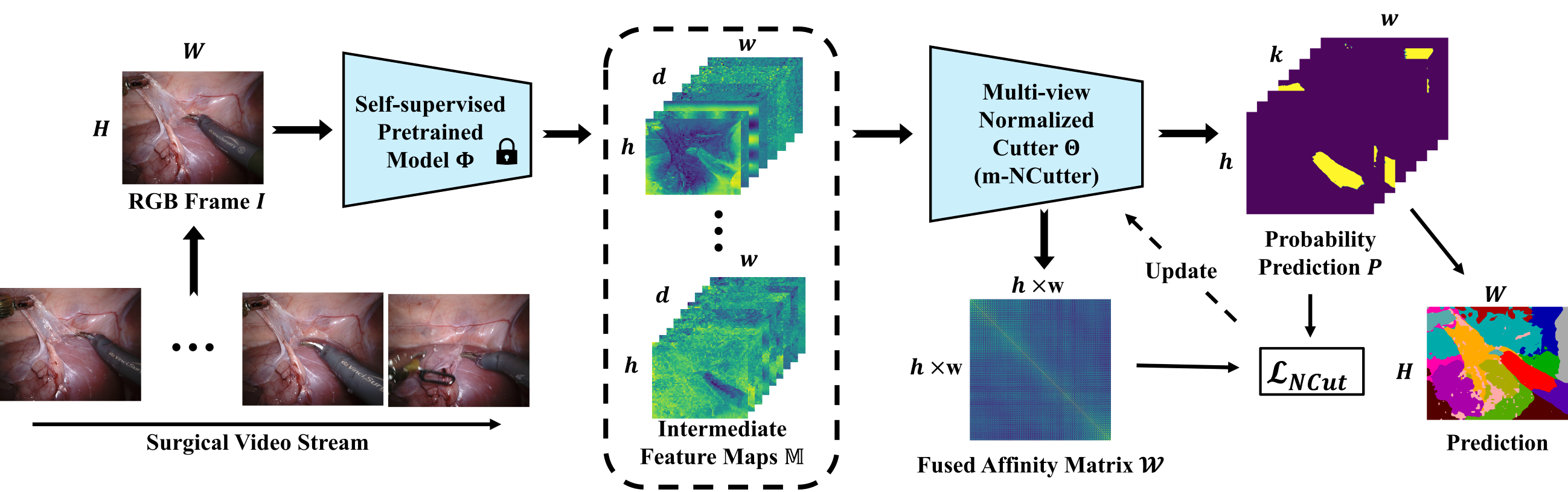}

   \caption{
    Method Overview.
   a) \textit{Pre-trained Backbone} introduced in \cref{Method: Backbone}; 
   b) \textit{m-NCutter}, our novel module presented in \cref{Method: MNCutter}; 
   c) \textit{NCut Loss} detailed in \cref{Method: mNCutLoss}. 
   }
   
    \label{Fig: method_overview}
\end{figure*}

\section{Related Work}
\label{RelatedWork}

        SIS aims to detect various surgical tools and patient organs. SIS tasks typically include binary, part, type, and semantic segmentation, serving different purposes: "binary" for distinguishing instruments from the tissue background, "part" and "type" for further identifying surgical tools' components and categories, respectively, and "semantic" for identifying all objects within the surgical video frame, including patient tissues and organs.
        

        \noindent
        \textbf{Supervised.} CNN is one of the most prevalent architectures for SIS because of its outstanding capability of extracting spatial features from images \cite{laina2017concurrent, garcia2017real, attia2017surgical, garcia2017toolnet, shvets2018automatic, milletari2018cfcm, pakhomov2019deep}. 
        Recent efforts have focused on improving model performance through three key aspects: input features, cost function, and network structure. 
        Utilizing additional features (e.g., optical flow and extra frames) as model inputs is a straightforward yet effective strategy \cite{jin2019incorporating, wang2021efficient}, but it requires more demanded computational resources and reduces efficiency. 
        Involving other downstream tasks (e.g., tool localization and pose estimation) and designing corresponding cost functions as auxiliary objectives can offer additional supervision for models. This approach is termed "multi-task learning", enabling models to learn comprehensive patterns and produce accurate segmentation masks \cite{islam2019learning, qin2019surgical, hasan2021detection, sanchez2021scalable, psychogyios2022msdesis}. However, supplementary annotation efforts and complex model structures present two challenges.
        Implementing a feature fusion structure, through either summation or concatenation, mitigates the issue of the model overly focusing on high-level features while neglecting essential low-level details \cite{ni2019rasnet, ni2020attention, ni2022surginet, ceron2022real, LSKANet_Liu}. Its effectiveness inspires our work, and we further extend it to produce affinities' attention scores, highlighting valuable affinities for segmentation.
        
        \noindent
        \textbf{Unsupervised.} The lack of labeled surgical data remains a significant hindrance, affecting the performance and robustness of supervised approaches. Consequently, USIS has emerged as a promising solution, although it remains under-explored with comparatively few related works.
        Most USIS studies rely on the pseudo-labels (e.g., physical positions of robotic tools, colors, lightness, shape priors, and optical flow) \cite{da2019self, liu2020unsupervised, alexe2010object, sestini2023fun}. However, this strategy requires additional effort in the pre-crafting of pseudo-labels and may contribute to inaccurate training due to unreliable supervision (e.g., noises, light reflection, and abrupt camera shake) \cite{wu2024rethinkinglowqualityopticalflow}. It also demonstrates weak generalization ability in different surgical scenes because the pseudo-labels are generated based on certain rules which may be inapplicable to unconsidered scenarios that involve more complexities. Furthermore, most pseudo-label-based methods are typically restricted to the binary task due to the nature of binary-only pseudo-labels (e.g., optical flow). 

\section{Method}
\label{Method}
    In order to avoid the limitations of the pseudo-label technique, fully leverage the intermediate feature maps, and achieve an end-to-end framework for real-time running, we propose a new USIS method AMNCutter whose overview is shown in \cref{Fig: method_overview}. More details of its modules are introduced in the following subsections.

    \subsection{Backbone}
    \label{Method: Backbone}
        DINO \cite{caron2021emerging}, a self-supervised ViT-based model, is prevalently employed as an effective pre-trained backbone, whose efficacy in feature extraction has been well-established in prior studies \cite{simeoni2021localizing, melas2022deep, hamilton2022unsupervised, Li10204554ACSeg, wang2023cut}. Consequently, we adopt DINO as the image encoder in our method, denoted as $\Phi$.
        Let $X \in \mathbb{R}^{H \times W \times 3}$ represent an RGB frame from a surgical video stream, where $H$ and $W$ denote frame height and width, respectively. The set of intermediate feature maps extracted from $\Phi$ is:

        \begin{equation}
        \label{Equ: FeatureMapSet}
          \mathbb{M} = \Phi ( X ) = \{ M_1, M_2, ..., M_\iota, ..., M_\mathcal{B} \},
        \end{equation}
        where $M_\iota \in \mathbb{R}^{h \times w \times d}$ is the $\iota$-th feature map extracted from a specific layer in $\Phi$; $|\mathbb{M}| = \mathcal{B}$ represents the total number of feature maps; $h$, $w$, and $d$ are height, width and channel dimensions of the feature map, respectively.

    \begin{figure*}[htbp]
      \centering
       \includegraphics[width=0.9\linewidth]{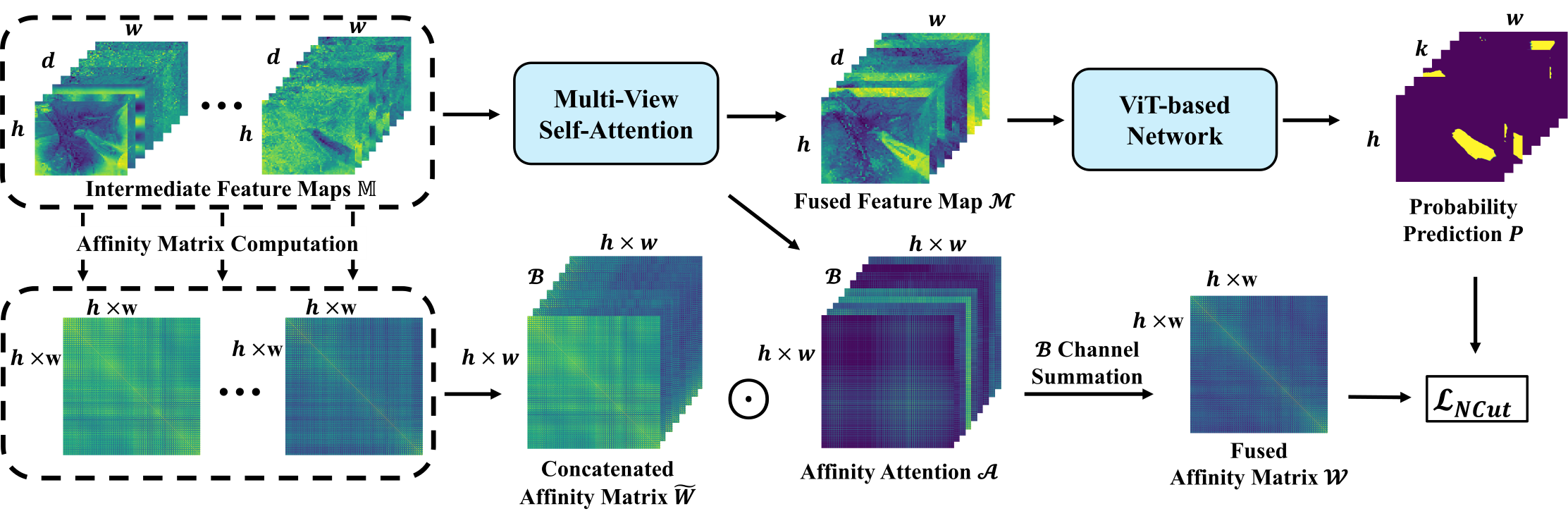}
    
       \caption{
          Multi-View Normalized Cutter (m-NCutter). The \textit{Multi-View Self-attention} is a key module, to produce the fused feature map and affinity-wise attention. "$\odot$": element-wise multiplication.
       }
       
       \label{Fig: MNcutterOverview}
    \end{figure*}
    
    \subsection{Multi-View Normalized Cutter}
    \label{Method: MNCutter}
        Let the m-NCutter be noted as $\Theta$, taking the intermediate feature maps as input and then returning probability prediction, which can be represented as:
        \begin{equation}
        \label{Equ: mNCutter}
          ( \mathcal{W}, P ) = \Theta ( \mathbb{M} ),
        \end{equation}
        where $\mathcal{W} \in \mathbb{R}^{s \times s}$ is a fused affinity matrix; $s = h \times w$ denote the number of patches; and $P \in \mathbb{R}^{h \times w \times k}$ indicates the probability prediction for the surgical frame $X$. The final segmentation mask $S \in \mathbb{R}^{H \times W}$ is obtained by resizing $P$ into $\tilde{P} \in \mathbb{R}^{H \times W \times k}$.

\begin{figure}[htbp]
      \centering
       \includegraphics[width=0.8\linewidth]{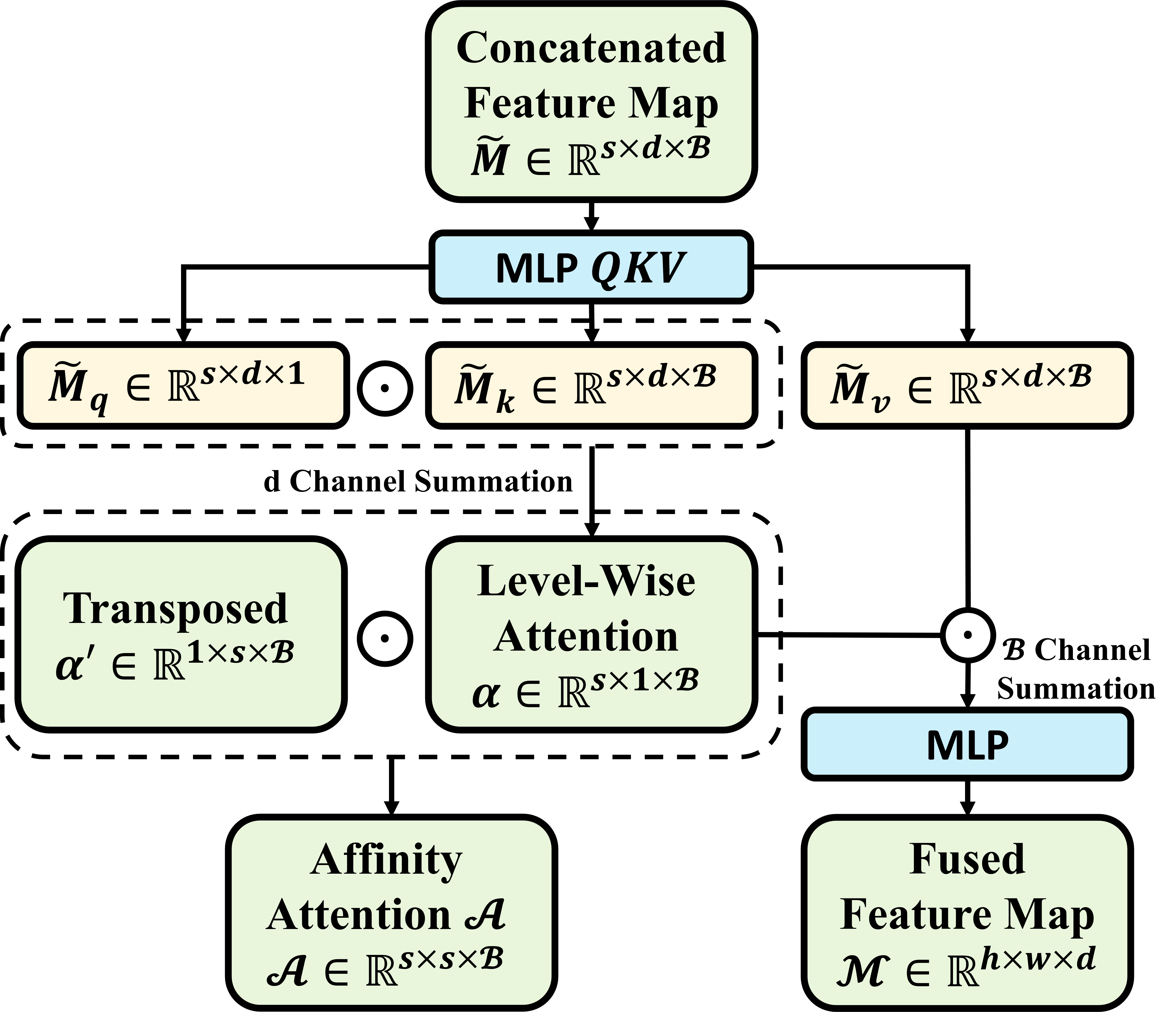}
    
       \caption{
       Multi-View Self-Attention. Only one head is shown for simplicity while using multi-heads in our experiment.
       }
       
       \label{Fig: MultiViewSelfAttn}
    \end{figure}

        The overview of the m-NCutter module is presented in \cref{Fig: MNcutterOverview}. First, the corresponding affinity matrices $W_\iota \in \mathbb{R}^{s\times s}$ are calculated from the feature maps $M_\iota$ by:

        \begin{equation}
        \label{Equ: AffinityMatrix}
          w_{i, j} = 
            \begin{cases}
              \text{cos}(f_i, f_j) & i \neq j \\
              0 & i = j
            \end{cases},        
        \end{equation}
        where "cos" denotes cosine similarity, $w_{i, j}$ represents affinity between patches $i$ and $j$; $f_i \in \mathbb{R}^{d}$ represents the feature vector for patch $i$; $w_{i, i} = 0$ follows \cite{Ncut}. Adaptively determining the magnitudes of affinities from different levels is a central aspect of our method, achieved through a novel \textit{Multi-View Self-attention} block (described in \cref{Method: MultiViewSelfAttn} and shown in \cref{Fig: MultiViewSelfAttn}). This block fuses feature maps $\mathbb{M}$ into $\mathcal{M} \in \mathbb{R}^{h \times w \times d}$ and yields affinity-wise attention scores $\mathcal{A} \in \mathbb{R}^{s \times s \times \mathcal{B}}$ for the affinity matrices. Let $\mathcal{A}_\iota \in \mathbb{R}^{s \times s}$ denote the affinity attention for $W_\iota$. The fused affinity matrix is computed as follows:

        \begin{equation}
            \label{Equ: AffinityMerge}
            \mathcal{W} = 0.5 \cdot \left( 1 + \sum_{\iota = 1}^{\mathcal{B}} W_\iota \odot \mathcal{A}_\iota \right),
        \end{equation}
        where the affinities are standardized into the range $[0, 1]$ to ensure the affinity matrix is positive semi-definite \cite{Ncut}.
        
        $\mathcal{M}$ is subsequently fed into a ViT-based network followed by a softmax layer to compute the probability prediction. We utilize the ViT-based network because of its superior capacity to learn and capture meaningful representations of patches and affinity features, analogous to node and edge features in graph theory, owing to its self-attention mechanism \cite{joshi2019efficientgraphconvolutionalnetwork, Vijay_JMLR_v24_22_0567, NEURIPS2020_94aef384, NEURIPS2020_94aef384}. Let $\sigma$ denote the number of attention blocks in the ViT-based network, which is evaluated with different values in our ablation study (\cref{ExpDis: AblationStudy}).

    \subsection{Multi-View Self-attention}
    \label{Method: MultiViewSelfAttn}

    Building on the original self-attention mechanism, we design a novel multi-view self-attention block (as shown in \cref{Fig: MultiViewSelfAttn}), to generate the fused feature map $\mathcal{M}$ and affinity-wise attention scores. 
    The matrices of query $\tilde{M}_q \in \mathbb{R}^{s \times d \times 1}$, key $\tilde{M}_k \in \mathbb{R}^{s \times d \times \mathcal{B}}$, and value $\tilde{M}_v \in \mathbb{R}^{s \times d \times \mathcal{B}}$ are derived from the concatenated feature map $\tilde{M} \in \mathbb{R}^{s \times d \times \mathcal{B}}$. The affinity-wise attention $\mathcal{A}$ is calculated by element-wise multiplication of $\alpha \in \mathbb{R}^{s \times 1 \times \mathcal{B}}$ and its transpose $\alpha' \in \mathbb{R}^{1 \times s \times \mathcal{B}}$. This approach allows the significance of an affinity between two patches to be governed by their attention scores. For example, if two patches have both high attention scores at a particular level, their affinity will be prioritized over others. The effectiveness of this module is justified through the ablation study in \cref{ExpDis: AblationStudy}.

    \subsection{Normalized Cut Loss}
    \label{Method: mNCutLoss}
        \noindent
        \textbf{Graph Cut.} Let $G=(V, E)$ denote a graph with node set $V$ and edge set $E$, and let $W \in \mathbb{R}^{s \times s}$ represent a positive semi-definite affinity matrix whose element $w_{i, j}$ reflects the affinity between nodes $i$ and $j$, where $s$ is the number of nodes. The cost to cut a graph into two groups, $A$ and $B$, is calculated as follows: 
        
        \begin{equation}
        \label{Equ: CutCost}
            Cut(A, B) = \sum_{i \in A} \sum_{j \in B} w_{i,j}.
        \end{equation}
        The formula sums the affinities between all pairs of node $i$ in cluster $A$ and $j$ in cluster $B$, thereby measuring the total weight of the edges removed or cut. Minimizing this cut cost yields the optimal bi-partitioning of the graph.
        
        \noindent
        \textbf{Normalized Cut.} For tackling the imbalance clustering issue, \cite{Ncut} advanced the basic graph cut cost into a Normalized Cut (NCut) cost, defined as:

        \begin{align}
        \begin{split}
        \label{Equ: NCutCost}
            NCut(A, B) &= \frac{Cut(A, B)}{Cut(A, V)} + \frac{Cut(A, B)}{Cut(B, V)} \\
                       &= \frac{S_A^T W S_B}{S_A^T W \textbf{1}} + \frac{S_A^T W S_B}{S_B^T W \textbf{1}} \\
                       &= 2 - \frac{S_A^T W S_A}{S_A^T W \textbf{1}} - \frac{S_B^T W S_B}{S_B^T W \textbf{1}},
        \end{split}
        \end{align}
        where $\textbf{1} \in \mathbb{R}^s$ is a vector of ones; $S_A$ and $S_B \in \mathbb{R}^s$ are two class indicators for clusters $A$ and $B$, respectively.
        \cite{Ncut} solved this optimization problem by rearranging and relaxing the \cref{Equ: NCutCost} into a Rayleigh quotient. The relaxed optimal solution is then obtained from the second smallest eigenvector of the Laplacian matrix:
        
        \begin{equation}
        \label{Equ: LaplacianMatrix}
            L = D^{\frac{1}{2}} (D - W) D^{\frac{1}{2}},
        \end{equation}
        where $D$ is a degree matrix, a diagonal matrix whose elements are row-summation of the affinity matrix $W$. 

        \noindent
        \textbf{Normalized Cut Loss.} The NCut Loss is an unsupervised loss function, representing a soft version of NCut, optimized through gradient descent. The effectiveness of the graph-cutting approach has been demonstrated in numerous studies \cite{NcutLoss, LiMultiNcutLoss, YIN2023212039, Aflalo10350524DeepCut, Fan_2024_Seeing}. The NCut Loss for bi-partitioning is defined as follows:

        \begin{equation}
            \label{Equ: NCutLoss}
            \mathcal{L}_\text{NCut} = 2 - \frac{P_A^T W P_A}{P_A^T W \textbf{1}} - \frac{P_B^T W P_B}{P_B^T W \textbf{1}},
        \end{equation}
        where $P_A$ and $P_B \in \mathbb{R}^s$ indicate probability vectors, a soft version of the class indicators $S$, for clusters $A$ and $B$.

        \noindent
        \textbf{Ours.} Different from previous works, our NCut Loss is calculated using the fused affinity matrix $\mathcal{W}$ and the multi-class probability prediction $P$, where $P$ is reshaped into $(s, k)$. Therefore, our model can handle multi-class unsupervised segmentation and integrate multi-level affinities, fully considering the patch affinities at different levels. The formula of our NCut Loss is written as:
        \begin{equation}
            \label{Equ: mNCutLoss}
            \mathcal{L}_\text{NCut} = 1 - \frac{1}{k} \sum_{c \in \mathcal{K}} \frac{P_c^T \mathcal{W} P_c}{P_c^T \mathcal{W} \textbf{1}},
        \end{equation}
        where $\mathcal{K}$ denotes the set of clusters labels; $k = |\mathcal{K}|$ is the total number of clusters; $c$ is a cluster label; $P_c \in \mathbb{R}^{s}$ is a probability vector of cluster $c$. With the affinity-attention-guided fused affinity matrix $\mathcal{W}$, our NCut Loss can comprehensively consider the nodes' correlation across various levels, thereby producing relatively more accurate results, which is compared with solely using the deepest-level affinities in our ablation study in \cref{ExpDis: AblationStudy}.

\begin{figure}[htbp]
      \centering
       \includegraphics[width=0.8\linewidth]{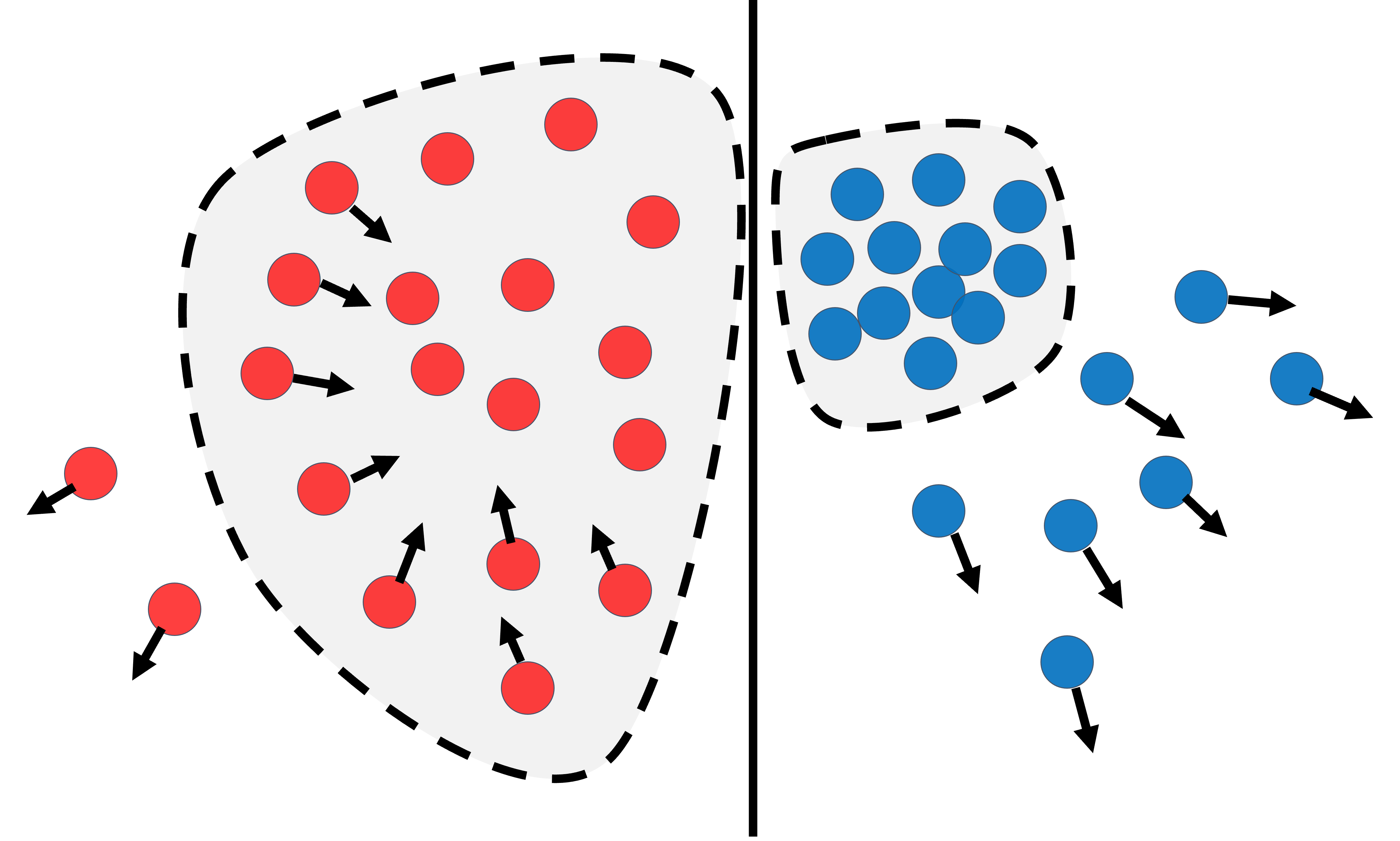}
    
       \caption{
       Different Effects on Loose and Tight Clusters. The loose cluster (left) attracts more nodes/patches due to its lower tightness $\tau_c$, whereas the tight cluster (right) repels nodes with lower probability because of its higher $\tau_c$. The arrow indicates "pull" or "push". For aesthetic purposes, not all arrows are drawn.
       }
       
       \label{Fig: ncut_loss_pull_push}
    \end{figure}

        \noindent
        \textbf{Gradient Computation.} In previous works, the derivative of the loss function $\mathcal{L}_\text{NCut}$, with respect to a cluster probability vector $P_c$, is calculated by:
        
        \begin{align}
        \begin{split}
        \label{Equ: NCutLossGradient1}
            \frac{\partial \mathcal{L}_{NCut}}{\partial P_c} 
            &= \frac{\mathcal{W} \textbf{1} \cdot P_c^T \mathcal{W} P_c - 2 \mathcal{W} P_c \cdot P_c^T \mathcal{W} \textbf{1}}
                    {\left( P_c^T \mathcal{W} \textbf{1} \right)^2}, 
        \end{split}
        \end{align}
        which demonstrates an unintuitive gradient. To delve deeper and facilitate a clearer understanding, we further calculate the derivative with respect to the probability $p_{i, c}$ of a patch $i$:

        \begin{align}
        \begin{split}
        \label{Equ: NCutLossGradient2}
            \frac{\partial \mathcal{L}_{NCut}}{\partial p_{i, c}} 
            &= \frac{\mathcal{W}_i^T \textbf{1}}{P_c^T \mathcal{W} \textbf{1}} \cdot \left( \frac{P_c^T \mathcal{W} P_c}{P_c^T \mathcal{W} \textbf{1}} - 2 p_{i, c} \right) \\
            &= \frac{\mathcal{W}_i^T \textbf{1}}{\textbf{1}^T \mathcal{W} \textbf{1}} \cdot \frac{\textbf{1}^T \mathcal{W} \textbf{1}}{P_c^T \mathcal{W} \textbf{1}} \cdot (\tau_c - 2 p_{i,c}) \\
            &= \gamma_{i} \cdot \frac{1}{\eta_{c}} \cdot (\tau_c - 2 p_{i,c}),
        \end{split}
        \end{align}
        where $\mathcal{W}_i \in \mathbb{R}^{s}$ is the $i$-th column of $\mathcal{W}$, indicating all affinities associated with patch $i$; $\tau_c = P_c^T \mathcal{W} P_c / P_c^T \mathcal{W} \textbf{1}$ measures tightness of cluster $c$; $\gamma_{i} = \mathcal{W}_i^T \textbf{1} / \textbf{1}^T \mathcal{W} \textbf{1}$ is the normalized degree for patch/node $i$, reflecting its hub-ness within the entire graph; similarly, $\eta_{c} = P_c^T \mathcal{W} \textbf{1} / \textbf{1}^T \mathcal{W} \textbf{1}$ is the normalized degree of the cluster/sub-graph $c$, reflecting the cost to separate the cluster from the whole graph \cite{Ncut, NcutLoss}. Obviously, with respect to $p_{i,c}$, the gradient magnitude is primarily influenced by $\gamma_{i}$ and $\eta_{c}$. High hub-ness patches and isolated clusters receive more gradients (simple-first). The sign of the gradient, whether positive or negative, is only determined by $\tau_c$ and $p_{i, c}$. For instance, for a tight cluster with high tightness $\tau_c$, patches with high probabilities ($p_{i, c} > 0.5 \tau_c$) gain negative gradients, which further increases their probabilities, pulling together, whereas patches with lower probabilities are pushed away. As a result, a tight cluster tends to exclude low-probability patches, while a loose cluster tends to attract patches. An example is provided in \cref{Fig: ncut_loss_pull_push}, assuming no interaction between the two clusters for simplicity.
        
\section{Experiments and Results}
\label{ExperimentDiscussion}

    \subsection{Experimental Setup}
    \label{ExpDis: ExperimentalSetup}

        \begin{table}[htbp]
        \centering
        \resizebox{1.0\linewidth}{!}{
            \begin{tabular}{l c c c c c}
        
                \toprule 
                Dataset & Binary & Part & Type & Semantic\\ 
                \midrule 
                EndoVis2017 \cite{allan20192017} & $\surd$ & 5 & 8 & $\times$ \\
                EndoVis2018 \cite{allan20202018} & $\surd$ & 4 & 7 & 12 \\
                ARTNetDataset \cite{hasan2021detection} & $\surd$ & $\times$ & $\times$ & $\times$ \\
                UCL \cite{colleoni2020synthetic} & $\surd$ & $\times$ & $\times$ & $\times$ \\
                CholecSeg8k \cite{hong2020cholecseg8k} & $\times$ & $\times$ & $\times$ & 13 \\
                \bottomrule 
            \end{tabular}
            }

        \caption{Datasets and Tasks. $\times$ means the task is officially unavailable. The numbers indicate the number of classes.}
        \label{Tab: Datasets}
        \end{table}

    \begin{table*}[htbp]
    \centering
    \resizebox{1.0\linewidth}{!}{
    
        \begin{tabular}{l | l c c c | c c c c c}
    
            \toprule 
            Method & Backbone & LF & E2E & MT & EndoVis2017 & EndoVis2018 & ARTNetDataset & UCL & All \\ 
            \midrule 
            FUN-SIS \cite{sestini2023fun} $^{*}$ & CycleGAN & $\times$ & $\surd$ & $\times$ & 76.25 $\pm$ 18.61 & - & - & - & - \\
            AGSD \cite{liu2020unsupervised} & UNet & $\times$ & $\surd$ & $\times$ & 83.65 $\pm$ 0.63 & 69.31 $\pm$ 4.25 & - & 80.54 $\pm$ 1.26 & 77.83 $\pm$ 6.16 \\
            DFC \cite{Kim9151332DFC} & CNN & $\times$ & $\surd$ & $\surd$ & 66.04 $\pm$ 0.90 & 56.38 $\pm$ 3.20 & 61.39 $\pm$ 4.54 & 63.33 $\pm$ 1.84 & 61.78 $\pm$ 3.53 \\
            GP-SIS \cite{sheng2024revisitingsurgicalinstrumentsegmentation} & DINO & $\surd$ & $\times$ & $\surd$ & 81.30 $\pm$ \textbf{0.15} & 80.15 $\pm$ \textbf{0.08} & 84.62 $\pm$ \textbf{0.26} & 80.90 $\pm$ \textbf{0.18} & 81.74 $\pm$ 1.71 \\
            STEGO \cite{hamilton2022unsupervised} & DINO & $\surd$ & $\times$ & $\surd$ & 77.65 $\pm$ 1.21 & 79.19 $\pm$ 2.28 & 85.37 $\pm$ 0.53 & 78.57 $\pm$ 2.04 & 80.19 $\pm$ 3.04 \\
            \midrule 
            Ours & DINO & $\surd$ & $\surd$ & $\surd$ & \textbf{84.78} $\pm$ 2.03 & \textbf{83.00} $\pm$ 1.86 & \textbf{85.48} $\pm$ 2.77 & \textbf{83.86} $\pm$ 1.53 & \textbf{84.28} $\pm$ \textbf{0.94} \\
            \bottomrule 
        \end{tabular}
        }

    \caption{
    Unsupervised Methods Comparison on the Binary Segmentation Task. "LF", "E2E" and "MT" represent "label-free", "end-to-end" and "multi-class tasks available", respectively. The measures on datasets are indicated with "Avg. $\pm$ Std." mIoU (\%), calculated from five different random seeds. The "All" column is calculated with mean and standard deviation across different scenes. \textbf{Bold} indicates the best across different methods. Methods noted with "$*$" are not reproduced, and their results are referenced from their reports. 
    }
    \label{Tab: BinaryComparison}
    \end{table*}
    
        \noindent
        \textbf{Datasets.} Extensive experiments are conducted across the datasets: EndoVis2017 \cite{allan20192017},  EndoVis2018 \cite{allan20202018}, ARTNetDataset \cite{hasan2021detection}, CholecSeg8k \cite{hong2020cholecseg8k}, and UCL \cite{colleoni2020synthetic}, as summarized in \cref{Tab: Datasets}. The datasets possess different characteristics. EndoVis2017 and EndoVis2018 datasets feature comparatively high-quality frames with a variety of surgical tools, some of which are very small and indistinguishable. The non-video-based ARTNetDataset offers relatively less training data. The synthetic UCL dataset combines ex vivo surgical tool foregrounds with animal tissue backgrounds. CholecSeg8k presents a narrow field of view within surgical video frames, often surrounded by extensive black edges. These datasets significantly challenge the model's performance and adaptability.

        \noindent
        \textbf{Implement Details.} In our experiment, for all methods: \textbf{1) ground-truth labels and testing data are not involved in training}; 2) ours and reproduced SOTA models are trained on the same device, a single NVIDIA Geforce RTX 3090, for 5 epochs (following \cite{liu2020unsupervised}) with a batch size of 4, using Adam optimizer configured with the same learning rate, beta1, and beta2; 3) all label-free methods are configured with the same number of clusters; 4) label matching between prediction and ground truth is applied to each prediction mask since label-free methods do not use label-based supervision; 5) data augmentation (e.g., flipping and cropping) and prediction post-processing like Conditional Random Field (CRF) are not employed, ensuring a fair comparison. Our model has 25.63M parameters and runs at approximately 240 FPS in testing, excluding data loading and label matching processes. 
        For more details and to reproduce our work, please visit our code.   

    \subsection{Comparison Results with SOTA Methods}
    \label{ExpDis: ComparisonResultsWithSOTAMethods}

        \begin{table}[htbp]
        \centering
        \resizebox{1.0\linewidth}{!}{
            \begin{tabular}{l | c c c}
        
                \toprule 
                Method & EndoVis2017 & EndoVis2018 & All \\ 
                \midrule 
                DFC \cite{Kim9151332DFC} & 38.64 $\pm$ \textbf{0.32} & 30.29 $\pm$ 2.63 & 34.47 $\pm$ 4.18 \\
                GP-SIS \cite{sheng2024revisitingsurgicalinstrumentsegmentation} & \underline{59.41} $\pm$ \underline{0.13} & \underline{58.33} $\pm$ \underline{0.10} & \underline{58.87} $\pm$ \textbf{0.54} \\
                STEGO \cite{hamilton2022unsupervised} & 46.21 $\pm$ 2.36 & 45.57 $\pm$ 2.37 & 45.89 $\pm$ \underline{0.32} \\
                Ours & \textbf{56.26} $\pm$ 2.22 & \textbf{54.57} $\pm$ \textbf{1.53} & \textbf{55.42} $\pm$ 0.85 \\
                \bottomrule 
            \end{tabular}
            }

        \caption{
        Part Segmentation Task (mIoU [\%]). \underline{Underline} and \textbf{bold} mark the best and second-best for each task.
        }
        \label{Tab: PartComparison}
        \end{table}

        \begin{table}[htbp]
        \centering
        \resizebox{1.0\linewidth}{!}{
            \begin{tabular}{l | c c c}
        
                \toprule 
                Method & EndoVis2017 & EndoVis2018 & All \\ 
                \midrule 
                DFC \cite{Kim9151332DFC} & 35.70 $\pm$ \textbf{0.48} & 21.43 $\pm$ 1.22 & 28.56 $\pm$ 7.13 \\
                GP-SIS \cite{sheng2024revisitingsurgicalinstrumentsegmentation} & \underline{58.86} $\pm$ \underline{0.24} & \underline{44.64} $\pm$ \textbf{0.35} & \underline{51.75} $\pm$ \textbf{7.11} \\
                STEGO \cite{hamilton2022unsupervised} & 44.56 $\pm$ 2.21 & 31.68 $\pm$ \underline{0.31} & 38.12 $\pm$ \underline{6.44} \\
                Ours & \textbf{56.01} $\pm$ 3.13 & \textbf{36.95} $\pm$ 1.64 & \textbf{46.48} $\pm$ 9.53 \\
                \bottomrule 
            \end{tabular}
            }

        \caption{
        Type Segmentation Task (mIoU [\%]).
        }
        \label{Tab: TypeComparison}
        \end{table}

        \begin{table}[htbp]
        \centering
        \resizebox{1.0\linewidth}{!}{
            \begin{tabular}{l | c c c}
        
                \toprule 
                Method & CholecSeg8k & EndoVis2018 & All \\ 
                \midrule 
                DFC \cite{Kim9151332DFC} & 34.45 $\pm$ 3.24 & 18.28 $\pm$ 1.47 & 26.37 $\pm$ \underline{8.08} \\
                GP-SIS \cite{sheng2024revisitingsurgicalinstrumentsegmentation} & \underline{63.69} $\pm$ \underline{0.07} & \underline{46.48} $\pm$ \underline{0.25} & \underline{55.08} $\pm$ 8.61 \\
                STEGO \cite{hamilton2022unsupervised} & 44.39 $\pm$ 2.40 & 27.92 $\pm$ \textbf{1.18} & 36.16 $\pm$ \textbf{8.23} \\
                Ours & \textbf{60.34} $\pm$ \textbf{2.31} & \textbf{39.46} $\pm$ 1.27 & \textbf{49.90} $\pm$ 10.44 \\
                \bottomrule 
            \end{tabular}
            }

        \caption{
        Semantic Segmentation Task (mIoU [\%]). 
        }
        \label{Tab: SemanticComparison}
        \end{table}

        \begin{figure*}[htbp]
        \centering    
           \includegraphics[width=1.0\linewidth]{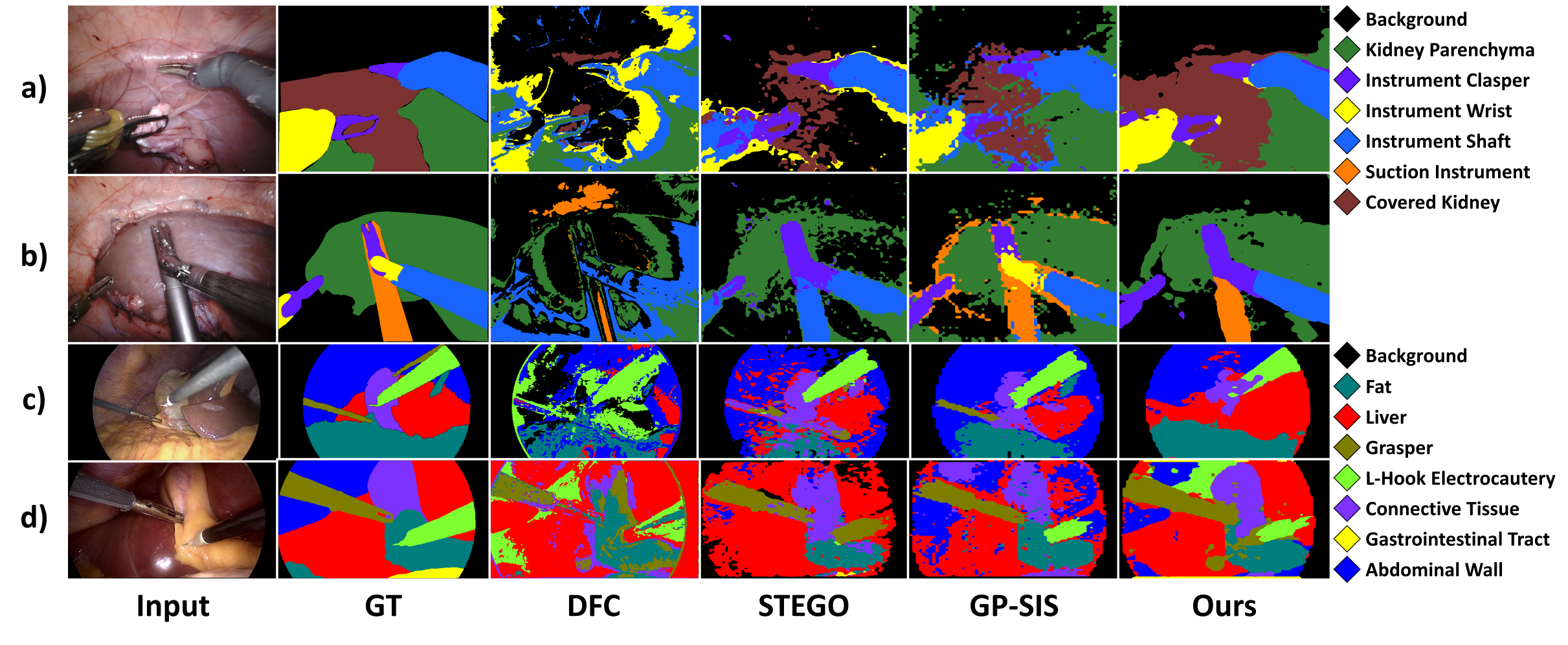}
        
           \caption{
           Semantic Segmentation Visualization. "a, b" and "c, d" are from the EndoVis2018 and CholecSeg8k datasets, respectively.
           }
           
            \label{Fig: semanticCompare}
        \end{figure*}

        \begin{figure}[htbp]
        \centering    
           \includegraphics[width=1.0\linewidth]{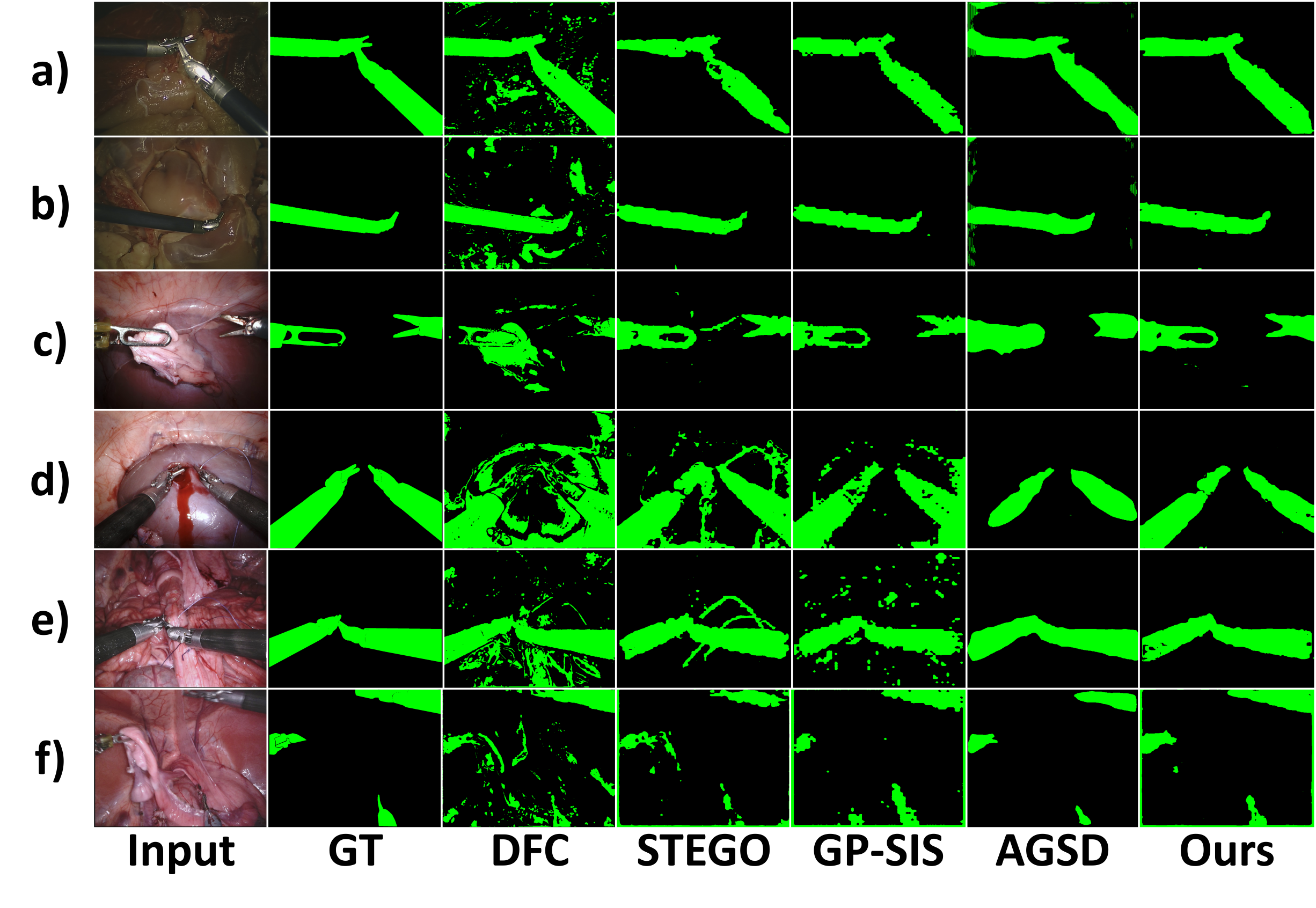}
        
           \caption{
           Binary Prediction Visualization. "a, b", "c, d" and "e, f" are from the UCL, EndoVis2018, and EndoVis2017 datasets, respectively.
           }
           
            \label{Fig: binaryCompare}
        \end{figure}

        \noindent
        \textbf{Binary Segmentation.} We evaluate our model on four distinct datasets: EndoVis2017, EndoVis2018, ARTNetDataset, and UCL. The results, reported in \cref{Tab: BinaryComparison}, demonstrate that our method outperforms (about + 3\% $\sim$ 23\% mIoU) other SOTA methods, exhibiting excellent robustness across all datasets (low Std. at $\pm$ 0.94\%). GP-SIS shows the least sensitivity to random seed variations, with a low Std. of $\pm$ 0.16\%, while the slowest computation (1.5 FPS). In terms of the generalization ability across different datasets, our method is slightly more robust than other label-free methods (i.e., DFC, GP-SIS, and STEGO), and significantly better than pseudo-label-based methods like AGSD, owing to the nature of label-free training and the limitation of pseudo-label learning. For instance, on the EndoVis2018 dataset, a relatively harder dataset involving more complexities than EndoVis2017, the AGSD suffers from severe performance degeneration, from 83.65\% on EndoVis2017 to 69.31\% on EndoVis2018, due to the complexities like severe light reflection. It poses difficulties for AGSD's pseudo-label generation process of AGSD, which primarily relies on colors.

        \noindent
        \textbf{Multi-Class Segmentation.} Results for part, type, and semantic segmentation tasks are reported in \cref{Tab: PartComparison,Tab: TypeComparison,Tab: SemanticComparison}. In the more challenging multi-class tasks, all methods exhibit comparatively lower mIoU than in the simple binary task. Our approach achieves the second-best mIoU across different datasets, higher than others by about 10\% $\sim$ 20\%. GP-SIS, while delivering the best performance, is the slowest running at 1.5 FPS and is unavailable for real-time running. Regrettably, for difficult tasks (i.e., type and semantic) involving a greater number of object classes, the robustness of our approach is slightly reduced. Despite the minor drawback, our method surpasses other SOTA approaches when considering both average accuracy and running efficiency. 


        \noindent
        \textbf{Results Visualization.} 
        To ensure a fair comparison and accurately showcase the model outputs, no post-processing technique is applied to the prediction masks, such as erosion, dilation, or CRF. Binary segmentation results of various methods are displayed in \cref{Fig: binaryCompare}. Our method yields relatively high-quality and precise results across various datasets. The pseudo-label-based method AGSD also produces high-quality results; however, some details of surgical instruments, such as the tool heads in \cref{Fig: binaryCompare}.c and tool shafts in \cref{Fig: binaryCompare}.d, are ambiguous due to inaccurate pseudo-labels. Other label-free-based methods (i.e., STEGO and GP-SIS) sometimes produce low-quality results with severe misdiagnoses, including misrecognized suturing threads in \cref{Fig: binaryCompare}.c-e. Qualitative results for multi-class (semantic) segmentation are shown in \cref{Fig: semanticCompare}, where our method demonstrates higher accuracy and quality. Our approach detects clearer kidneys and instruments (\cref{Fig: binaryCompare}.a and b). Tissues and organs (e.g., fat and liver) are segmented with comparatively greater completeness, as shown in \cref{Fig: binaryCompare}.c and d.

    \subsection{Ablation Study}
    \label{ExpDis: AblationStudy}
    
        \begin{table*}[htbp]
        \centering
        \resizebox{1.0\linewidth}{!}{
            \begin{tabular}{c | c c c | c c c | c c c | c c c}
        
                \toprule 
                 ~ & \multicolumn{3}{c|}{Binary Segmentation Task} & \multicolumn{3}{c|}{Part Segmentation Task} & \multicolumn{3}{c|}{Type Segmentation Task} & \multicolumn{3}{c}{Semantic Segmentation Task} \\ 
                \midrule 
                \diagbox[height=1.5\line]{$k$}{$\sigma$} & 1 & 3 & 6 $\uparrow$ & 1 & 3 & 6 $\uparrow$ & 1 & 3 & 6 $\uparrow$ & 1 & 3 & 6 $\uparrow$ \\ 
                \midrule 
                5 & 72.68 / \textbf{73.14} & 74.82 / \textbf{77.03} & \textbf{78.41} / 77.03 & \textbf{46.66} / 43.48 & 44.65 / \textbf{45.55} & 44.43 / \textbf{45.79} & 27.84 / \textbf{28.56} & 28.30 / \textbf{28.64} & \textbf{29.86} / 29.15 & \textbf{29.73} / 29.18 & 29.76 / \textbf{30.49} & 29.64 / \textbf{30.52} \\
                10 & \textbf{82.85} / 81.70 & \textbf{83.55} / 83.47 & \textbf{82.66} / 81.45 & \textbf{55.48} / 51.64 & \textbf{55.26} / 53.82 & \textbf{54.26} / 52.13 & \textbf{36.87} / 34.27 & \textbf{34.53} / 34.44 & \textbf{34.35} / 34.22 & \textbf{37.40} / 35.12 & \textbf{37.17} / 36.10 & \textbf{36.63} / 35.78 \\
                15 & 81.16 / \textbf{81.43} & 79.21 / \textbf{79.70} & \textbf{82.60} / 78.69 & 52.66 / \textbf{54.75} & 50.77 / \textbf{52.23} & \textbf{55.28} / 51.70 & 36.01 / \textbf{36.62} & 33.86 / \textbf{34.41} & \textbf{37.30} / 34.16 & 37.70 / \textbf{38.70} & 34.12 / \textbf{37.05} & \textbf{40.03} / 36.89 \\
                20 & 80.95 / \textbf{82.57} & \textbf{81.24} / 80.78 & \textbf{83.94} / 80.40 & 53.83 / \textbf{55.47} & 54.48 / \textbf{56.27} & \textbf{57.70} / 55.25 & 35.64 / \textbf{39.13} & \textbf{37.89} / 36.90 & \textbf{39.62} / 37.21 & 39.40 / \textbf{40.28} & \textbf{41.21} / 40.31 & \textbf{42.85} / 41.19 \\
                30 & \underline{\textbf{85.15}} / 82.08 & \textbf{84.82} / 81.45 & \textbf{84.41} / 80.04 & 56.97 / \textbf{58.02} & \underline{\textbf{59.68}} / 56.27 & \textbf{58.29} / 55.15 & \textbf{39.50} / 38.41 & \underline{\textbf{41.27}} / 40.46 & \textbf{40.55} / 39.94 & \textbf{43.05} / 41.75 & \textbf{44.41} / 43.54 & \underline{\textbf{44.75}} / 42.32 \\
                \bottomrule 
            \end{tabular}
            }

        \caption{
        Ablation Study for All Segmentation Tasks on EndoVis2018 (mIoU [\%]). "- / -" indicates the comparison of performance between multi-view NCutter (m-NCutter, left) and non-multi-view one (NCutter, right), where the NCutter only learns from the deepest feature map and its affinity matrix; the higher value is marked in \textbf{bold}. $k$ represents the number of clusters. $\sigma$ represents the number of attention blocks for the ViT-based Network. \underline{Underline} marks the best across all $k$ and $\sigma$ for each task. "$\uparrow$" highlights a $\sigma$ value where our novel multi-view mechanism fully demonstrates its superior performance.
        }
        \label{Tab: AblationStudy}
        \end{table*}

        We conduct ablation experiments to justify the effectiveness of our multi-view mechanism by using the EndoVis2018 dataset, which encompasses all four segmentation tasks. The results are reported in \cref{Tab: AblationStudy}. Our m-NCutter demonstrates superior performance with higher values of $\sigma$ (e.g., $\sigma$ = 6). The parameter $\sigma$ reflects the model's complexity and learning ability, indicating that our multi-view affinity fusion strategy is more effective for complex models. Simpler models may struggle to effectively process the intricate and extensive affinity features generated by the multi-view self-attention block. 
        Regarding the number of clusters, increasing $k$ generally improves performance due to over segmentation. For each task, our m-NCutter consistently achieves superior performance, as emphasized by the underlined results. Additionally, the optimal $\sigma$ values vary between simple and complex tasks, with simpler tasks benefiting from a shallower network and more challenging tasks requiring a deeper network that is overqualified for the simple tasks.

    \subsection{Transfer Learning}
    \label{ExpDis: TransferLearning}
    
        Our approach demonstrates outstanding capability as a pre-trained model. We extend our pre-trained AMNCutter model by adding several convolutional layers as a segmentation head behind the m-NCutter module. We evaluate our method on the EndoVis2017 dataset, which is widely used in existent supervised SIS works \cite{gonzalez2020isinet, ni2020attention, islam2019learning, jin2019incorporating, pakhomov2019deep, wang2021efficient, shen2023branch, Yue_Zhang_Hu_Xia_Luo_Wang_2024}. To ensure a fair comparison: our model is pre-trained exclusively on the EndoVis2017 dataset, without access to extra datasets; only the segmentation head is trained with ground-truth labels; the mIoU results are measured over five tests using different random seeds; each test leverages a pre-trained AMNCutter model initialized with the corresponding seed.  

        \begin{table}[htbp]
        \centering
        \resizebox{1.0\linewidth}{!}{
            \begin{tabular}{l | c c c}
        
                \toprule 
                Method & Binary & Part & Type \\ 
                \midrule 
                BAANet \cite{shen2023branch} & - & - & \underline{61.59} $\pm$ 9.40 \\
                AGLN \cite{ni2020attention} & - & - & 58.30 $\pm$ 26.95 \\
                SurgicalSAM \cite{Yue_Zhang_Hu_Xia_Luo_Wang_2024} & - & - & \textbf{67.03} $\pm$ \; \, / \; \;  \\
                DMNet \cite{wang2021efficient} & - & - & 53.89 $\pm$ \; \, / \; \; \\
                ISINet \cite{gonzalez2020isinet} & - & - & 38.08 $\pm$ \; \, / \; \; \\
                LWTL \cite{islam2019learning} & 87.56 $\pm$ 16.24 & 67.92 $\pm$ 16.50 & 36.62 $\pm$ 22.78 \\
                MF-TAPNet \cite{jin2019incorporating} & 87.56 $\pm$ 16.24 & 67.92 $\pm$ 16.50 & 36.62 $\pm$ 22.78 \\
                DRLIS \cite{pakhomov2019deep} & 89.60 $\pm$ 2.92 & \textbf{76.40} $\pm$ 9.16 & - \\
                \midrule 
                Ours + CNN$_1$ & 90.95 $\pm$ \underline{0.48} & 64.59 $\pm$ 0.96 & 40.02 $\pm$ 1.46 \\
                Ours + CNN$_3$ & \textbf{91.87} $\pm$ \textbf{0.40} & 68.64 $\pm$ \underline{0.88} & 42.38 $\pm$ \textbf{1.10} \\
                Ours + CNN$_5$ & \underline{91.85} $\pm$ 0.66 & \underline{70.72} $\pm$ \textbf{0.87} & 43.27 $\pm$ \underline{1.19} \\
                \bottomrule 
            \end{tabular}
            }

        \caption{
        Supervised Methods on the EndoVis2017 Dataset (mIoU [\%]). Other supervised SIS methods' results are referenced from their study reports. "-" and "/" mark unprovided results. \textbf{Bold} and \underline{underline} mark the best and second-best for each task. "CNN$_i$" indicates a segmentation head comprising $i$ convolutional layers.
        }
        \label{Tab: TransferLearningSup}
        \end{table}
        
        The results are reported in \cref{Tab: TransferLearningSup}. For the binary task, our method achieves the highest mIoU score (91.87\%) with a low Std. ($\pm$ 0.40\%) across five different random seeds, where our model is extended with a simple segmentation head consisting of three convolutional layers. 
        For the part task, "Ours + CNN$_5$" ranks second with a mIoU of 70.72\%, falling short of the best results by 5.70\%; for the type task, it shows relatively lower performance (43.27\%).
        Our method shows the most superior performance in the binary segmentation task. For multi-class segmentation tasks, although our approach does not achieve SOTA performance, it shows an upward trend in performance as the complexity of the segmentation head increases. Notably, our method consistently maintains the highest stability (low Std. around 1.00\%) across all tasks. This experiment confirms the remarkable potential of our approach for use as a pre-trained model in transfer learning.

\section{Conclusion}
\label{Conclusion}
    
    This work proposes a fully label-free USIS method, named AMNCutter. Unlike other SOTA USIS methods, our approach does not depend on pseudo-labels, demonstrating exceptional generalization capability. We devise a novel module, the m-NCutter, which adaptively captures multi-level features and investigates their affinities at different levels. These affinities are weighted using affinity attention scores derived from a novel Multi-View Self-Attention module, guiding the model in determining which affinities from which levels are meaningful or irrelevant. The model is trained with NCut Loss, a graph-cutting loss inspired by the NCut method, with affinities among patches serving as supervision. Extensive and rigorous experiments across various datasets confirm our method's superior performance, robustness, and potential as a pre-trained model. We conduct ablation studies to verify the effectiveness of m-NCutter and suggest a proper hyper-parameter setting. However, due to the characteristics of the backbone model (DINO) and computational resource limitations, the raw predictions are low-resolution (patch-wise) rather than high-resolution (pixel-wise), which may lead to coarse detection and reduced segmentation quality. Therefore, future works may focus on generating high-resolution masks for more precise object detection. 
    
{
\small
\bibliographystyle{ieee_fullname}
\bibliography{myBiB}
}

\end{document}